\documentclass[conference]{ieeeconf}
\IEEEoverridecommandlockouts
\usepackage{cite}
\usepackage{amsmath,amssymb,amsfonts}
\usepackage{algorithmic}
\usepackage{graphicx}
\usepackage{textcomp}
\usepackage{fancyvrb}
\usepackage{multirow}
\let\labelindent\relax
\usepackage{enumitem}
\usepackage{booktabs}
\usepackage{listings}
\usepackage{tcolorbox}
\usepackage{tabularx}
\usepackage{array}
\usepackage{url} 
\usepackage[edges]{forest}
\usepackage{hyperref}

\providecommand{\IEEEauthorblockN}[1]{#1}
\providecommand{\IEEEauthorblockA}[1]{\\ #1}

\tcbuselibrary{listings, skins}

\def\BibTeX{{\rm B\kern-.05em{\sc i\kern-.025em b}\kern-.08em
    T\kern-.1667em\lower.7ex\hbox{E}\kern-.125emX}}

\begin{document}

\title{
Declarative Reactive Robot Behaviors from Imperative Language Models
}

\author{
\IEEEauthorblockN{Vincent Berenz\textsuperscript{1},
Tasha Kim\textsuperscript{2},
Bernhard Schölkopf\textsuperscript{1,3,4}}\\
\IEEEauthorblockA{\textsuperscript{1}Max Planck Institute for Intelligent Systems, Tübingen, Germany\\
\textsuperscript{2}University of Oxford, Oxford, United Kingdom\\
\textsuperscript{3}ELLIS Institute Tübingen, Germany \quad
\textsuperscript{4}ETH Zürich, Zürich, Switzerland\\
\{vincent.berenz, bernhard.schoelkopf\}@tuebingen.mpg.de,\ tasha.kim@eng.ox.ac.uk}
\thanks{\copyright~2026 IEEE. Personal use of this material is permitted. Permission from IEEE must be obtained for all other uses, in any current or future media, including reprinting/republishing this material for advertising or promotional purposes, creating new collective works, for resale or redistribution to servers or lists, or reuse of any copyrighted component of this work in other works. Accepted to the 2026 IEEE International Conference on Robot and Human Interactive Communication (RO-MAN).}
}

\maketitle

\begin{abstract}

Large Language Models (LLMs) are predominantly trained on imperative programming languages, and existing work on LLM-based robot programming reflects this: it focuses on goal-directed tasks expressed as plans. However, many robotic applications require not tasks but \emph{behaviors}--open-ended, continuous patterns of sensory-motor engagement with no termination condition (e.g., greeting visitors, patrolling an area). Such behaviors are better served by declarative and reactive programming, on which LLMs have virtually no training. We investigate whether this gap can be bridged. Using \textsc{Playful}, a reactive scripting language, we show that an LLM-based coding agent can generate correct declarative and reactive programs for robotic behaviors. These programs are produced from under-specified natural language descriptions, given suitable context and a set of reusable programming guidelines. The guidelines encode design principles that redirect the agent from imperative defaults toward proper reactive patterns. Developed once by an expert, they generalize across robots and behaviors. Starting from only the generic robot operating system API, the agent jointly designs both the imperative component (Python modules for sensing, acting, and evaluating) and the declarative component (the reactive script that composes them into behavior logic). The declarative nature of reactive programming supports both incremental refinement through conversational turns and one-shot generation of complete behaviors. %

\end{abstract}

\section{Introduction}

The vision of users programming robots through natural conversation is a longstanding goal of human-robot interaction research~\cite{tellex2020robotslanguage, ajaykumar2022eupsurvey, gorostiza2011dialog}. Recent advances in speech recognition and LLMs have made this increasingly feasible. A common pipeline transcribes speech, interprets it using an LLM, then converts it into executable code~\cite{wang2025llm_robotics_survey, cohen2024grounding_survey}. 
Existing work has largely focused on \textit{task-oriented} programming, which entails generating programs that achieve a specific goal with explicit success criteria and termination conditions (e.g., ``bring me a coffee''). Such tasks naturally align with imperative programming, since the structure of plan execution can be modeled by sequential control flow, explicit state transitions, and conditional branching. This is precisely what LLMs excel at, as they are trained predominantly on imperative programming languages. 

In this paper, we address the distinct challenge of programming robotic \textit{behaviors}, a term we use to denote open-ended, continuous patterns of sensorimotor engagement that persist and adapt over time (e.g., ``greet visitors in a friendly manner''), which are not reduced to a single success criterion or termination condition. This kind of persistent reactivity has long been studied in rule-based agent architectures and robot control systems~\cite{brooks1986layered, nilsson1994teleoreactive}. In contemporary robotics, behavior trees (BTs) are often used to organize reactive behavior logic~\cite{colledanchise2018btbook, iovino2022btsurvey}. Programming behaviors requires specifying \textit{what} a robot should do under \textit{what conditions}, with continuous evaluation and adaptation, which are requirements that are naturally captured by declarative and reactive programming~\cite{berenz2014tdm, berenz2018playful}. Yet, LLMs trained predominantly on mainstream imperative languages have very limited exposure to declarative or reactive paradigms, particularly in the context of domain-specific languages (DSLs)~\cite{joel2024dsl_survey}.

This raises the question of whether LLMs can generate correct declarative and reactive programs despite this training gap. We investigate whether this gap can be bridged using \textsc{Playful}~\cite{berenz2018playful}, a behavior programming platform with two components. The first is an \textit{imperative} component: Python modules implementing sensing, acting, and evaluation primitives that interface with the robot's operating system. The second is a \textit{declarative and reactive} component: a scripting language composing these modules into hierarchical condition-action mappings that are continuously evaluated at runtime to form BTs~\cite{colledanchise2018btbook}. The declarative script encodes the logic of the behavior, and the Python modules provide the interface to the robot. 

A further challenge is that natural language behavior descriptions are inherently \textit{under-specified}: users rely on common-sense assumptions, rarely detail low-level aspects of the desired behavior (e.g., at what distance to stop, where to direct gaze, how to arbitrate competing actions), and may leave parts intentionally open. BTs are well suited to this problem: the top levels of the tree correspond to what the user \emph{does} specify, while the lower levels encode details the user leaves implicit. We ask if an LLM can fill in these lower levels appropriately, and whether the development process allows the user to observe and correct mismatches incrementally. We investigate whether, given suitable context (language documentation, a developer tutorial, and a set of reusable programming guidelines) an LLM-based coding agent can generate correct \textsc{Playful} programs from under-specified descriptions, despite having no prior training on the language or the reactive paradigm. A key empirical finding motivates this study: without explicit paradigm-level guidelines, the agent produces syntactically valid but poorly structured programs, in which behavioral logic leaks into Python procedural code rather than being expressed through the DSL's declarative mechanisms. This is a direct consequence of the agent's imperative defaults. But when provided with guidelines authored once by an expert, the same agent reliably produces well-structured reactive programs. This demonstrates that LLMs \emph{can} work with declarative paradigms when given sufficient context that illustrates how the language works and what correct design looks like. The declarative nature of \textsc{Playful} also supports incremental conversational refinement~\cite{iovino2022btsurvey}: since statements are independent and order-irrelevant, the coding agent can add or modify rules based on conversational turns without restructuring existing code. 

The agent starts from only the generic robot operating system (OS) API. Then it leverages its native strength in imperative code where it applies, to jointly design both the imperative Python modules and the declarative script that composes them into behavior logic. The guidelines steer it through the declarative component. This co-design produces a reusable primitive library that transfers across qualitatively different behaviors, with the agent inventing domain-appropriate primitives as needed. We evaluate this approach on the Pepper humanoid robot~\cite{pandey2018pepper} with two qualitatively different behaviors across three experiments: step-by-step conversational development, one-shot generation of the same behavior, and one-shot generation of a novel behavior. %

\section{Background and Related Work}

\subsection{End-user and natural-language robot programming} Programming robots through natural interaction is a longstanding challenge in human-robot interaction~\cite{ajaykumar2022eupsurvey}. Early dialog-based systems allowed end users to define robot social behaviors through structured conversation~\cite{gorostiza2011dialog}. 
The present work contributes to this tradition: by pairing an LLM coding agent with a declarative reactive language, we enable behavior specification through natural language, with the generated script serving as a transparent, human-readable artifact that users can inspect and refine. This is a property that purely neural approaches do not provide.

\subsection{LLMs for task-oriented robot execution} LLMs have been extensively applied to task planning and execution. Frameworks such as SayCan~\cite{ahn2022saycan} and Inner Monologue~\cite{huang2022innermonologue} ground LLM outputs in feasibility and affordance estimates, with execution feedback enabling replanning. ProgPrompt~\cite{singh2023progprompt} structures task planning as program synthesis using situated context and pre-conditions. LLMs have also been used to generate planning domain specifications directly~\cite{oswald2024domain_generators}. These systems excel at decomposing commands into skill sequences and adapting upon failure, but their outputs are fundamentally \textit{episodic}: structured around achieving a specified objective and terminating, rather than maintaining persistent reactive engagement with the environment. A complementary direction trains generalist visuomotor policies end-to-end ~\cite{brohan2022rt1}~\cite{zitkovich2023rt2}~\cite{driess2023palme}~\cite{bousmalis2023robocat}~\cite{octo2024}, but these yield opaque policies that users cannot inspect or incrementally refine--a contrast with the transparent, editable scripts produced here.%

\subsection{LLM code generation for robotics} LLMs are predominantly trained on imperative languages, reflected by code generation approaches for robotics; `Code as Policies'~\cite{liang2022codeaspolicies} synthesizes programs calling perception and control APIs; VoxPoser~\cite{huang2023voxposer} generates code constructing 3D value maps for manipulation; `Language to Rewards'~\cite{yu2023l2r} authors reward functions for reinforcement learning. Surveys of robotic language grounding~\cite{cohen2024grounding_survey} and LLM applications in robotics~\cite{wang2025llm_robotics_survey} confirm that generated artifacts are predominantly imperative: they encode a specific control flow toward task completion. The produced code is editable, but are not naturally \textit{compositional}.
These approaches also assume a suitable API and generate code that \emph{calls} it, but do not generate the API-level implementation itself. LLMs have also been applied to generate programs in domain-specific languages where training data is scarce~\cite{joel2024dsl_survey}, which is central to our work: \textsc{Playful} is a reactive DSL with minimal representation in standard LLM training corpora, motivating use of explicit documentation and expert-authored guidelines as context. %

 \subsection{LLMs and behavior trees} Several recent works have explored LLM-based BT generation in robotics. BTGenBot~\cite{izzo2024btgenbot} and BTGenBot-2~\cite{izzo2026btgenbot2} fine-tune small language models to generate BTs for task-oriented robot behaviors from natural language. BT-ACTION~\cite{leszczynski2025btaction} uses a test-driven approach to decompose user instructions into modular BTs via LLMs. A code-driven approach~\cite{zhang2025codebt} generates BTs from natural language using programmatic intermediate representations. Styrud et al.~\cite{styrud2026betrgui} present an assisted programming interface combining user interaction with LLM-based BT generation. BTs have also been learned from demonstration~\cite{french2019bt_lfd, scherf2023ilbert}, and combined with teleoperation interfaces for collaborative programming~\cite{paxton2017costar}. These works share our interest in LLM-based BT generation, but focus on \textit{task-oriented} behaviors in standard BT formalisms where node types and tree structure are predefined. In contrast, we target \textit{open-ended behaviors} using \textsc{Playful}'s reactive DSL, and require the agent to jointly design both the declarative BT \emph{and} the imperative Python modules that implement its primitive nodes, which is a co-design problem not addressed in prior BT generation work.

\subsection{Declarative and reactive behavior programming} Brooks' subsumption architecture~\cite{brooks1986layered} and Nilsson's teleo-reactive programs~\cite{nilsson1994teleoreactive} established the foundations of rule-based, layered reactive robot control. Formal approaches to reactive synthesis from natural language include temporal logic frameworks: LTLMoP~\cite{finucane2010ltlmop} maps natural language to linear temporal logic  for robot control synthesis, while recent LLM-based autoformalization approaches translates it to temporal or first-order logic specifications~\cite{chen2023nl2tl,nl2fol}; recent work applies grammar-forcing~\cite{english2025graft} or code synthesis~\cite{rabiei2025ltlcodegen} to improve syntactic correctness of generated specifications. These approaches give formally verifiable artifacts, but are oriented toward task-like safety and liveness constraints, not continuously evaluated, open-ended sensorimotor behaviors.

BTs have become a widely used formalism for hierarchical reactive behavior with runtime selection and fallback~\cite{colledanchise2018btbook, iovino2022btsurvey}. Targets-Drives-Means (TDM) proposed composing BTs via independent declarative statements and priority-based arbitration, with demonstrated usability advantages over traditional flow-chart methods~\cite{berenz2014tdm, berenz2012tdmbench}. \textsc{Playful} builds on TDM, implementing a reactive scripting language whose declarative statements resemble natural language~\cite{berenz2018playful}. Prior work has shown that \textsc{Playful} scripts can be generated from human demonstrations~\cite{berenz2020sensorymotor}. Our work targets generation from natural language alone, and unlike all prior \textsc{Playful} work, requires no pre-built primitive library.

\section{Reactive Programming and \textsc{Playful} Platform}
\label{sec:playful}

\begin{figure}[!ht]
  \centering
  \includegraphics[scale=0.09]{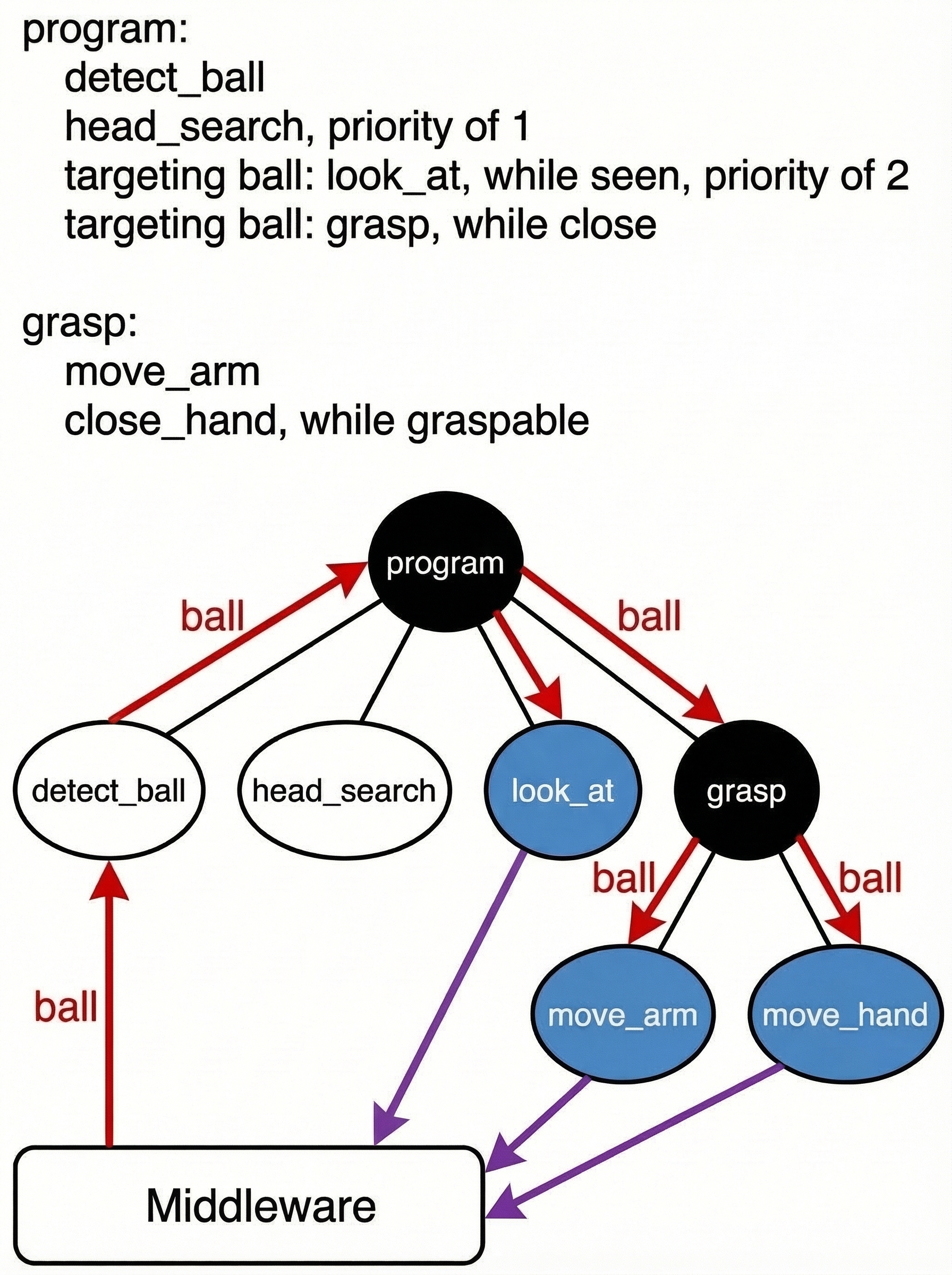}
  \caption{A \textsc{Playful} script and its behavior tree (BT) representation. The red and violet arrows explicit show how sensorimotor loops are enforced.}
  \label{fig:playful-tree}
\end{figure}

Consider a robot that should search for a ball, look at it when detected, and grasp it when close.
Figure~\ref{fig:playful-tree} displays the corresponding script as well as its graphical representations.

Rather than encoding this as a sequence of actions, \textsc{Playful} declares what conditions should trigger what actions. The order of statements is irrelevant: all the four statements of the program are continuously evaluated at runtime. Primitives (\texttt{detect\_ball}, \texttt{head\_search}, \texttt{look\_at}, \texttt{grasp}) are activated when their conditions (\texttt{seen}, \texttt{close}) hold. When primitives compete for a shared resource (e.g. \texttt{head\_search} and \texttt{look\_at} both control the head) the higher priority wins. The \emph{targeting} keyword binds a world-model object (\texttt{ball}) to the associated primitives, and the engine instantiates one set of nodes per detected object. If multiple objects are present, \emph{priority of} resolves conflicts via dynamic arbitration, continuously recomputed as the environment changes. Conditions, priorities, and leaf nodes are implemented as Python modules that interface with the robot's operating system. Scripts can reference other scripts hierarchically: for example, \texttt{grasp} is a `sub' \textsc{Playful} script combining \texttt{move\_arm} and \texttt{close\_hand}. The result is a behavior tree (Figure~\ref{fig:playful-tree}) whose branches are activated and deactivated at runtime based on evaluations and priorities.

A \textsc{Playful} application thus comprises two components: an \emph{imperative} component, reusable Python modules that interface with the robot OS; and a \emph{declarative} component: the script that composes these modules into behavior logic. A detailed description of the formal semantics is in~\cite{berenz2018playful}.

\section{Experiments}

\begin{figure}[t!]
  \vspace{6pt}
  \centering
  \includegraphics[width=\columnwidth]{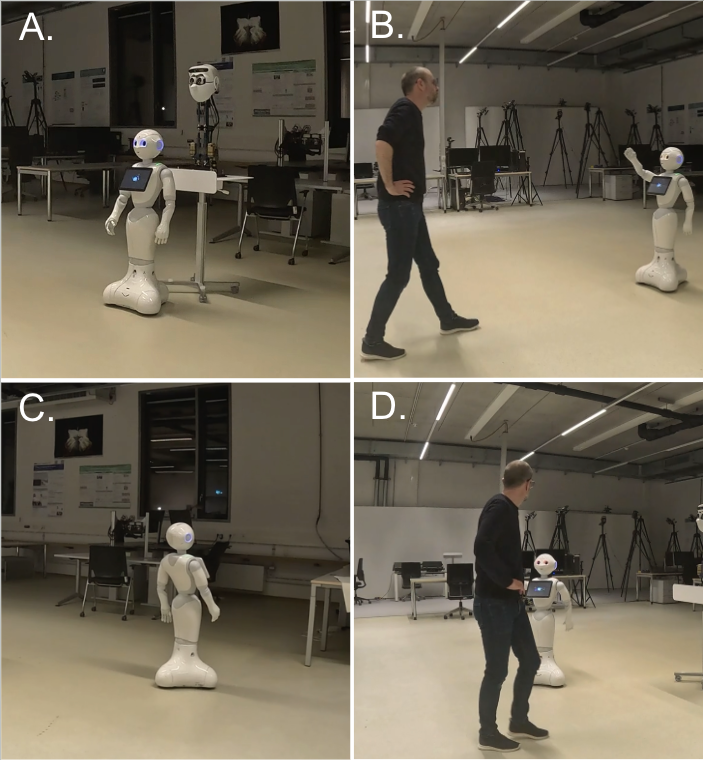}
  \caption{Experimental validation of the generated behaviors on Pepper. The \emph{greeter} behavior: Pepper robot transitions from idle scanning A. to active engagement B., driven by the declarative script using the targeting keyword. The \emph{guardian} behavior: Pepper robot patrols C. and intercepts a visitor with stop gesture D., governed by a script that dynamically arbitrates priority, and reuses primitives from the \emph{greeter} behavior.}
  \label{fig:eval-pepper}
\vspace{-1em}
\end{figure}

The experiments are designed to demonstrate that a coding agent without prior training on \textsc{Playful} can generate well-structured robotic applications when guided by paradigm-level design guidelines. Starting from underspecified natural language instructions, the agent jointly develops the imperative layer (Python modules) and the declarative layer (behavior trees). All experiments are conducted using the Pepper humanoid robot~\cite{pandey2018pepper} and Claude Code (Anthropic, 2025) Opus~4.6 as the coding agent. The context provided to the agent consists of:
\begin{itemize}
\item the \textsc{Playful} language description~\cite{berenz2018playful},
\item a \textsc{Playful} developer tutorial with illustrative examples, and
\item the Pepper robot's robot operating system (ROS) API reference.
\end{itemize}

The robot API follows a conventional Python-based robotics middleware design, wrapping NaoQi OS (SoftBank Robotics), ROS~\cite{quigley2009ros}, and OpenPose~\cite{cao2017openpose}. It is not specific to \textsc{Playful} and adheres to standard robotics interface conventions. Importantly, the API is intentionally limited in scope: it exposes only a set of basic, preemptible functions providing high-level access to sensors and actuators, and it is not integrated with advanced control or optimization libraries. This design choice ensures that the observed behavioral complexity arises from the proposed programming approach rather than from capabilities implicitly provided by the underlying middleware.
No pre-existing \textsc{Playful} primitive library is included in the agent’s context. Consequently, the agent must implement all Python artifacts--including scheme types, evaluation functions, and leaf nodes--alongside the behavior trees that orchestrate them.
The natural language instructions (Figure~\ref{fig:instructions}) define two behaviors, referred to as \emph{greeter} and \emph{guardian}. These instructions are deliberately under-specified; for instance, they do not explicitly require the robot to search for humans, prioritize battery management, or specify LED control policies. We provide the complete experimental context, including the guidelines (Section~\ref{sec:guidelines}), generated code, and execution videos online.\footnote{Code and videos available at: \url{https://nextcloud.tuebingen.mpg.de/index.php/s/qq932bWGDAip4Yd}}

\subsection{Developing the Guidelines}
\label{sec:guidelines}

\begin{figure}[t]
\rule{0pt}{0pt}%
\centering
\small
\begin{tcolorbox}[
  colback=white,
  colframe=black,
  boxrule=0pt,
  toprule=1.5pt,
  bottomrule=1.5pt,
  boxsep=0pt,
  colbacktitle=white,
  coltitle=black,
  fonttitle=\bfseries,
  arc=0pt,
  left=6pt, right=6pt, top=0pt, bottom=0pt,
  width=\dimexpr\columnwidth-2pt\relax,
  subtitle style={
    before skip=0pt,
    after skip=3pt,
    boxsep=2pt,
    top=1pt,
    bottom=1pt,
    left=4pt,
    right=4pt,
  },
]
\tcbsubtitle[before skip=0pt]{Behavior~1 (Greeter):}
\newcommand{\prompt}[2]{%
  \noindent\textbf{#1.}\quad\textit{``#2''}\par
}
\prompt{1}{Create this new behavior for the Pepper robot: the robot is at a designated location, looking for humans. When it sees one, it waves at them and goes toward them.}

\noindent\hspace*{1.6em}\textit{``The designated location and the charger do not correspond to physical objects that can be seen by the robot. They are imaginary locations tracked by odometry.''}\par\smallskip

\prompt{2}{The robot should never go too far from its designated location.}

\prompt{3}{When at its designated location, the robot should also turn toward the direction it had at startup.}

\prompt{4}{The robot should go to its charging station when its battery is low.}

\prompt{5}{While going to the charger or while charging, the robot should still interact with humans by turning toward them and waving at them.}

\prompt{6}{Manage the color of the LEDs to be entertaining and informative.}
\smallskip

\tcbsubtitle{Behavior~2 (Guardian):}

\prompt{1}{Given a location to guard, the robot should patrol around it and
(kindly) block people from approaching the location. The robot should find a
way to communicate with people that the location is off limits.}
\smallskip

\end{tcolorbox}
\caption{Natural language instructions that are provided to the coding agent. Behavior~1 was developed both in a single prompt (see Section~\ref{sec:behavior}),
and incrementally, with each step issued in a separate conversational turn (see
Section~\ref{sec:iterative}).}
\label{fig:instructions}
\end{figure}

\begin{table}[t]
\rule{0pt}{0pt}%
\caption{Programming guidelines provided to the coding agent}
\label{tab:guidelines}
\centering
\small
\begin{tabularx}{\columnwidth}{@{} p{2.0cm} >{\raggedright\arraybackslash}X @{}}
\toprule
\textbf{Category} & \textbf{Overview} \\
\midrule
Script as source of truth &
  The playful script must fully describe the behavior; reading Python
  should not be necessary. \\
\addlinespace[5pt]
Targeting and generality &
  Nodes and evaluations are generic; scheme-type binding is expressed
  only in the script via \texttt{targeting}. \\
\addlinespace[5pt]
Scheme types &
  Distinct behavioral roles require distinct scheme types. \\
\addlinespace[5pt]
Naming &
  Names must be generic and action-oriented. \\
\addlinespace[5pt]
Resource granularity &
  Resources map one-to-one to independently controllable API channels
  and to nodes. \\
\addlinespace[5pt]
Safety &
  Safety constraints must be visible in the script, not hidden in
  Python. \\
\addlinespace[5pt]
Oscillation &
  Resource arbitration is winner-takes-all with no blending; priority
  crossovers are discrete switches. \\
\addlinespace[5pt]
Code reuse &
  Reuse Python nodes, evaluations and priorities developed for other behaviors \\
\bottomrule
\end{tabularx}
\end{table}

During an exploratory phase, the coding agent was tasked with implementing various behaviors. Although the provided context was adequate for generating syntactically correct programs, the agent tended to default to imperative patterns. Specifically, behavioral logic was often implemented within lower-level Python modules rather than being expressed through the DSL operators (\emph{targeting}, \emph{while}, \emph{priority of}) that are designed to capture this logic at the script level. This approach risks obscuring the behavior specification, undermining the purpose of the reactive declarative formalism, and limiting code reusability. To address these recurring design issues, an expert user compiled a set of reusable, platform-independent guidelines (Table~\ref{tab:guidelines}). Their central objective is to make explicit to the coding agent that behavioral structure and arbitration must reside in the declarative script, while Python modules should remain minimal, reusable building blocks. The guidelines focus on enforcing a proper use of the \emph{targeting} mechanisms, and also address other aspects such as resource granularity and code reuse. Although these guidelines were developed in response to the agent's shortcomings, they were not derived from any systematic analysis of the agent's generated programs, which might have rendered them agent-specific. Instead, they remain generic: they would prove equally useful to human programmers.

\subsection{Developing the Behavior}
\label{sec:behavior}

\begin{figure}[!t]
  \centering
  \includegraphics[width=0.99\columnwidth]{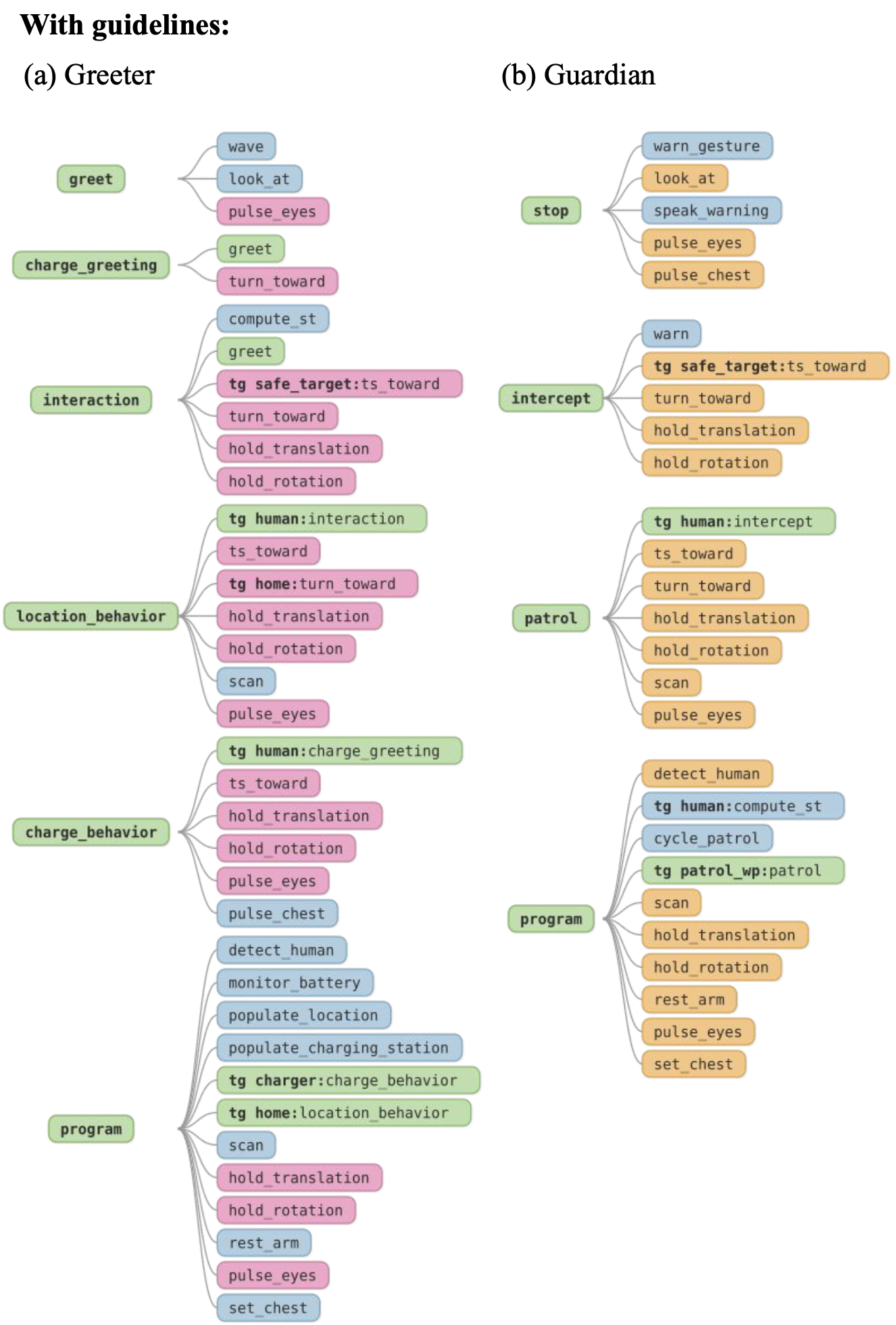}

  \vspace{1em}
  \includegraphics[width=0.92\columnwidth]{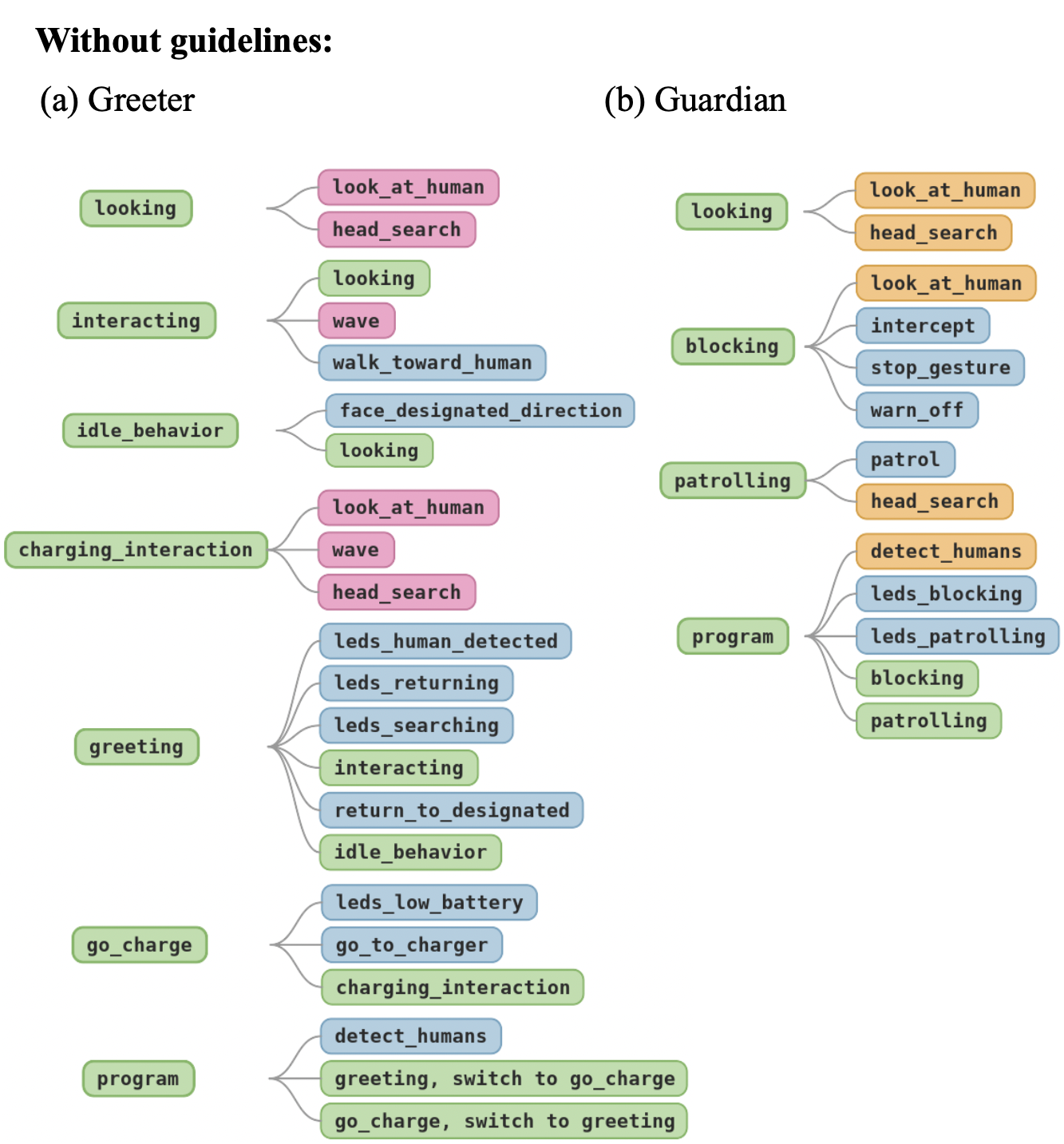}

    \vspace{0.5em}
  \caption{Flattened tree representations of the \textsc{Playful} scripts.}
  \label{fig:behavior-tree}
\end{figure}

The agent was instructed to implement the behaviors listed in Figure~\ref{fig:instructions}, both in the absence and in the presence of paradigm-level guidelines provided in the context. In both conditions, the agent generates complete and functional implementations of the greeter and guardian behaviors in a single pass, including both the Python modules and their orchestration through \textsc{Playful} scripts. The corresponding tree representations are shown in Figure~\ref{fig:behavior-tree}. In the figure, branches are displayed in green. `Program' is the top node. For the \emph{greeter} behavior, nodes which are reused in various branches are in pink. For the \emph{guardian} behavior, nodes reused from the \emph{greeter} behavior are in orange. Despite the under-specified nature of the natural language instructions--which do not explicitly require human search, battery management prioritization, LED control policies, or enforcement of a distance constraint from the home location--the agent infers reasonable default behaviors. As discussed in the following sections, however, the implementations produced with and without guidelines differ significantly in terms of script readability and code reusability.

\subsection{The Impact of Guidelines}
\label{sec:comparison}

\begin{table}[t]
\centering
\rule{0pt}{0pt}%
\caption{Distribution of logical elements between the Playful script and Python code across both behaviors}
\vspace{-0.5em}
\label{tab:logic_distribution}
\begin{tabular}{l r r}
\toprule
\textbf{Element type} & \textbf{Guidelines} & \textbf{No guidelines} \\
\midrule
\multicolumn{3}{l}{\textit{Script logical elements}} \\[2pt]
\quad Activation conditions (\texttt{targeting}) & 10 & 0 \\
\quad Activation conditions (\texttt{while}) & 10 & 16 \\
\quad Priority assignments & 56 & 20 \\
\quad Mode transitions & 0 & 4 \\
\quad \textit{Subtotal} & \textit{76} & \textit{40} \\
\multicolumn{3}{l}{\textit{Python logical elements}} \\[2pt]
\quad Memory reads & 3 & 11 \\
\quad Memory writes & 5 & 6 \\
\quad Behavioral conditions & 11 & 24 \\
\quad \textit{Subtotal} & \textit{19} & \textit{41} \\
\midrule
\textbf{Ratio (script\,/\,Python)} & \textbf{4.0} & \textbf{1.0} \\
\bottomrule
\end{tabular}
\end{table}

The primary structural difference between the two implementations lies in the use of the \emph{targeting} keyword, which appears exclusively in the guideline-compliant scripts. This construct makes explicit the relationships between objects, actions (nodes), and behavioral logic (evaluations), enabling reactive behaviors to be expressed declaratively rather than encoded imperatively. As a result, behavioral logic is shifted out of the Python layer and into the scripts.

This effect is reflected quantitatively in Table~\ref{tab:logic_distribution}. With guidelines, most logical elements are encoded in the declarative scripts, yielding a script-to-Python logic ratio of 4.0. Activation conditions and priority assignments are handled at the script level, while Python modules remain largely free of control logic. In contrast, without guidelines, activation conditions are implemented using imperative \emph{while} constructs, and a substantial fraction of behavioral conditions and memory accesses is embedded in Python, resulting in a balanced logic distribution (ratio 1.0).

The increased declarativity enabled by \emph{targeting} directly improves code reuse. \emph{Reused} lines are definitions from the greeter's Python file that the guardian script references directly. As shown in Table~\ref{tab:lines_reuse}, the guideline-compliant guardian behavior reuses 61\% of the Python code defined for the greeter, compared to 15\% without guidelines, despite similar total code size. This indicates that making object–action–logic relationships explicit promotes more generic and reusable Python primitives.

\begin{table}[t]
\rule{0pt}{0pt}%
\centering
\caption{Lines of code (Python) required for the guardian behavior}
\vspace{-0.5em}
\label{tab:lines_reuse}
\begin{tabular}{l r r}
\toprule
 \textbf{Code source} & \textbf{Guidelines} & \textbf{No guidelines} \\
\midrule
New lines (from guardian) & 144 & 263 \\
Reused lines (from greeter) & 225 & 47 \\
Total lines & 369 & 310 \\
\midrule
\textbf{Reuse ratio} & \textbf{61\%} & \textbf{15\%} \\
\bottomrule
\end{tabular}
\vspace{-1em}
\end{table}

\subsection{Iterative Development}
\label{sec:iterative}

The greeter behavior was also developed using the guidelines through six successive natural language instructions (Figure~\ref{fig:instructions}). At each iteration, the agent produced a functional behavior tree that was incrementally extended: new requirements were integrated into the existing script without requiring substantial restructuring. This demonstrates that the proposed approach supports iterative behavior development, enabling users to progressively refine and correct robot behavior through multiple conversational interactions.

\section{Discussion and Future Work}
\label{sec:discussion}

The experiments demonstrated that a coding agent, with no training on \textsc{Playful}, can: (1)~produce well-designed applications when guided by paradigm-level programming guidelines; (2)~jointly develop both the imperative layer (Python modules) and the declarative layer (behavior tree); and (3)~handle the under-specification inherent in natural language behavior descriptions.

A key finding of our work is that syntactic correctness and sound reactive design are distinct capabilities. When lacking guidelines, the agent produces valid but structurally incorrect programs. In particular, the behavioral logic migrated into Python instead of staying visible in the declarative script. This is a paradigm orientation failure, not of language understanding. That is, the agent's imperative defaults are strong enough to persist, even when DSL offers explicit declarative mechanisms. Guidelines communicate design intent without adding syntax, which redirects the agent toward patterns that it would not have independently selected. This suggests that for reactive or declarative DSL with limited training representation, a likely necessary component for LLM-based  paradigm-level guidelines of this kind is likely a necessary component for an LLM-based generation approach.

The experiments show that the agent can develop both layers from high-level natural-language descriptions, without prior training on the specific reactive paradigm: a tutorial and a set of guidelines included in the agent's context were sufficient. A notable outcome is that a reusable library of Python primitives (scheme types, evaluation functions, and leaf nodes) was co-developed alongside the scripts, and that these primitives transferred across behaviors.

The coding agent paired with a reactive formalism handles under-specification well. High-level instructions were compiled into behavior trees with multiple layers of abstraction. Much of the lower-level code (sensor processing, proportional control, LED patterns, derived scheme types) was only implicitly specified in the natural language instructions, yet was nevertheless implemented correctly. The co-developed Python primitive library transferred across qualitatively different behaviors, suggesting that a reusable layer accumulates naturally, even without being explicitly requested.
Another advantage of the proposed approach is that the resulting behavior is fully specified in a compact script that is easy to understand or amend. The resulting artifact is inspectable, and an expert user can easily audit and refine this.

Our experiments serve as an initial proof of concept that underscores several directions for further investigation. First, the robotic platform and API are intentionally constrained, and the deployed behaviors are deliberately simple, allowing the proposed approach to be examined in isolation; future work should extend the approach to more complex and modern robotic systems to assess scalability and generality. Second, future work could develop principled methods for authoring, refining, or automatically inducing guidelines, reducing reliance on expert knowledge. Lastly, the complete user-facing workflow should be evaluated, in which the embedded speech-to-text software feeds instructions to the agent that deploys the robot's behavior live; this evaluation would also include user-feedback studies on how well the generated behavior aligns with the user's expectations.

\section{Conclusion}
\label{sec:conclusion}

LLMs are trained predominantly on imperative code. On the other hand, robotic behaviors are often more naturally expressed through declarative and reactive programming. Our work shows this mismatch is not fundamental. That is, a set of paradigm-level guidelines that are authored once by an expert is sufficient to redirect a coding agent towards sound reactive design. Given these guidelines, a standard language documentation, and robot OS API, an LLM-based coding agent is able to jointly design both the imperative Python modules and the declarative reactive scripts. Through this, they produce deployable robotic applications from natural language alone.

Though the imperative defaults of an agent are strong, they remain unfixed and can be overriden through explicit design principles conveyed through context. This highlights a general strategy when applying LLM-based code generation to declarative and reactive DSLs, which are domains where training data is sparse, but structure of good design can be articulated. The resulting scripts are executable and readable, as they provide artifacts that users can inspect and refine through conversation. This turns reactive scripting into a productive interface that bridges natural language and robot control.

\bibliographystyle{ieeetr}
\bibliography{references}

\end{document}